%% file: main.tex
\documentclass[10pt,twocolumn,letterpaper]{article}

\usepackage[pagenumbers]{iccv} 

\input{preamble}

\definecolor{iccvblue}{rgb}{0.21,0.49,0.74}
\usepackage[pagebackref,breaklinks,colorlinks,allcolors=iccvblue]{hyperref}

\usepackage{xcolor}

\title{OmniPaint: Mastering Object-Oriented Editing via Disentangled Insertion-Removal Inpainting}

\author{%
   Yongsheng Yu\textsuperscript{1 *}, Ziyun Zeng\textsuperscript{1 *},  Haitian Zheng\textsuperscript{2}, Jiebo Luo\textsuperscript{1}\\
    \textsuperscript{1}University of Rochester, \textsuperscript{2}Adobe Research \\
    {\tt \small \{yyu90,zzeng24\}@ur.rochester.edu, hazheng@adobe.com, jluo@cs.rochester.edu} \\
 }

\begin{document}
\twocolumn[{%
\renewcommand\twocolumn[2][]{#1}%
\maketitle%
\vspace{-0.5cm}	
\input{sec/fig_teaser}\vspace{0.55cm}
}] 

\input{sec/0_abstract}    
\vspace{-0.5cm}

\section{Introduction}  

Object-oriented image editing has evolved from simple pixel-level adjustments to complex scene manipulation tasks, including object removal~\cite{ekin2024clipaway,powerpaint,yildirim2023inst,objectdrop} and insertion~\cite{addit,anydoor,paintbyexample}. Classic approaches for object removal/insertion in images have followed two distinct technical routes without intersection, such as object harmonization~\cite{azadi2020compositional,wu2019gp} and image completion~\cite{rombach2022high,li2022mat}. Recent advances in large diffusion-based generative models~\cite{rombach2022high,flux2024} have broadened the horizons of object-oriented editing, enabling not only high-fidelity inpainting of masked regions~\cite{powerpaint,FLUX-Inpainting,freecompose} but also creative synthesis of new objects seamlessly integrated into existing images~\cite{objectstitch,imprint,addit,anydoor}. These models further allow manual manipulation of object attributes and appearances through text prompts or reference images, demonstrating unique industrial value for visual content modification and creation.

Despite the transformative potential of diffusion-based models, their application to general object editing presents unique challenges. The first challenge lies in the path dependency on large-scale paired real-world datasets~\cite{objectdrop,objectmate} or synthetic datasets~\cite{jiang2025smarteraser,li2025rorem,magiceraser}. For specific tasks like object insertion, achieving correct geometric alignment and realistic integration requires not only high-quality synthesis but also deep understanding of complex physical effects like shadows, reflections, and occlusions. Insufficient paired training samples may lead to models lacking identity consistency or failing to integrate objects with realistic physical effects~\cite{objectdrop}. The second challenge involves ensuring reliable object removal that not only eliminates unwanted foreground elements but also maintains background continuity and prevents unintended introduction of artifacts or hallucinated objects~\cite{ekin2024clipaway} - particularly problematic given the lack of robust evaluation metrics for flagging ghost elements generated by large models' random hallucinations.

These limitations have led to separate modeling of object removal and insertion, whether text-driven~\cite{yildirim2023inst,DBLP:conf/iclr/FuHDWYG24,addit,yu2024promptfix} or mask-guided~\cite{powerpaint,ekin2024clipaway,imprint,objectstitch,objectdrop}. However, deploying large generative models simultaneously across different editing subtasks (e.g., removal and insertion) that currently employ completely different technical implementations risks potential conflicts and increased costs.

To address these challenges, we propose \textbf{OmniPaint} - a framework that reconceptualizes object removal and insertion as {\it interdependent tasks} rather than isolated subtasks. Leveraging pre-trained diffusion priors (employing FLUX~\cite{flux2024} in this work), we optimize LoRA~\cite{DBLP:conf/iclr/HuSWALWWC22} parameters through collected small-scale real-world paired samples while enabling easy task switching via learnable text embeddings. For realistic object removal, our model achieves semantic elimination of masked foreground elements while removing their physical effects. For object insertion, we go beyond simple blending to achieve harmonious synthesis respecting scene geometry and reference identity through our proposed CycleFlow mechanism. By {\it incorporating well-trained removal parameters into insertion training}, we enable the utilization of large-scale unpaired samples, significantly reducing dependence on massive real paired datasets.

\begin{figure}[t]
    \centering
    \includegraphics[width=\linewidth]{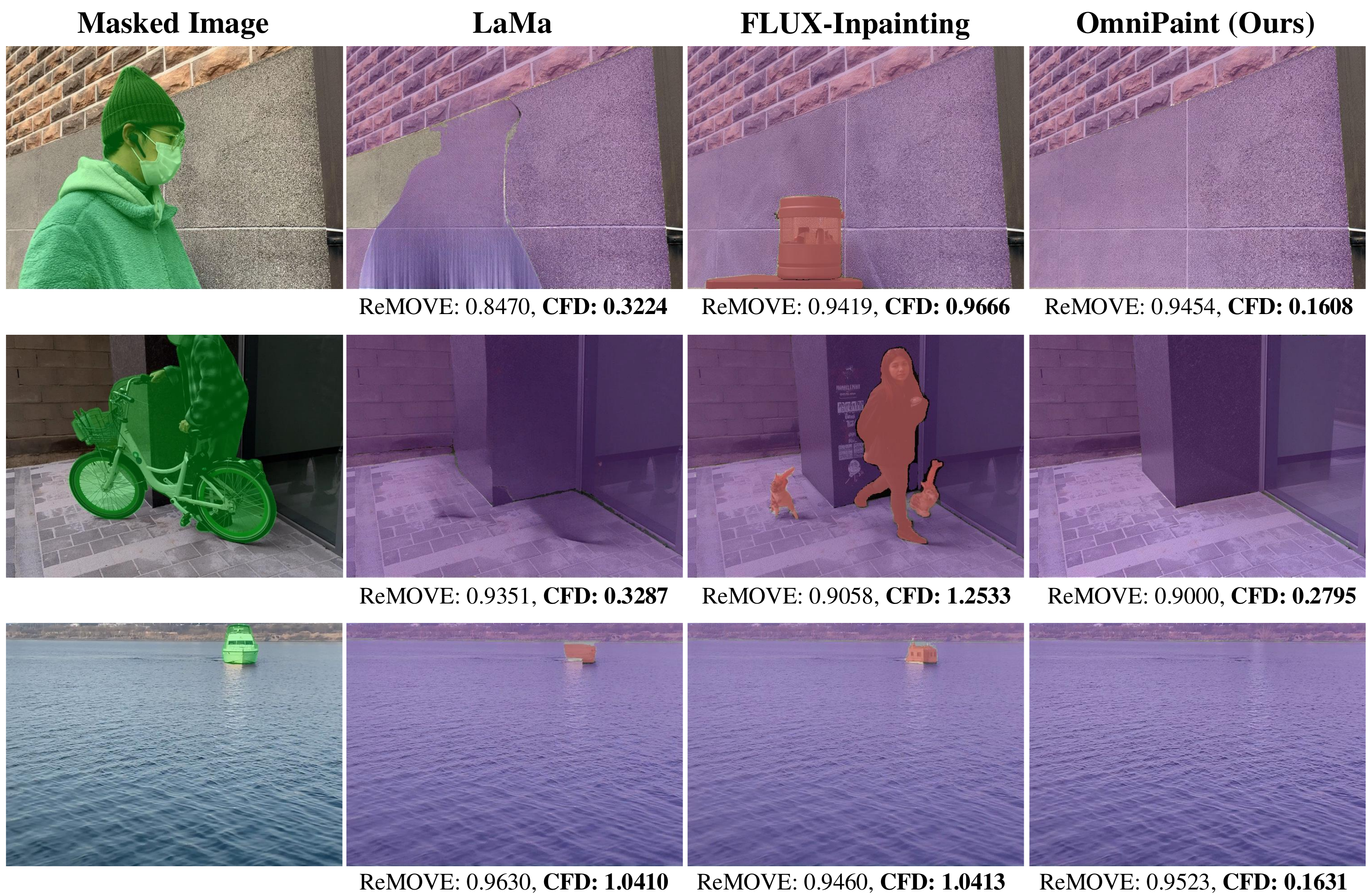}
    \caption{Visualization of CFD metric assessment for object removal. The segmentation results are obtained using SAM~\cite{sam1} with refinement, with purple masks for background, orange masks for segments fully within the original mask, and unmasked for those extending beyond the original mask. Note that the orange masked regions correspond to hallucinated objects. A higher ReMOVE~\cite{REMOVE} score is better, while a lower CFD score is preferable. In these cases, ReMOVE scores are too similar to indicate removal success, while CFD score offers a clearer distinction.}
    \vspace{-3mm}
    \label{fig:cfd_analysis}
\end{figure}

A key innovation is our Context-Aware Feature Derivation (CFD) score, a specialized no-reference metric for object removal. As illustrated in Fig.~\ref{fig:cfd_analysis}, it evaluates object hallucinations and context coherence, setting a new standard for realistic object-oriented editing. Experiments demonstrate OmniPaint's significant improvements in both interdependent editing tasks: the model better handles complex physical effects like shadows and reflections during removal while achieving seamless background reconstruction, and generates more natural geometric alignment and illumination consistency during insertion. Ablation studies reveal that omitting CycleFlow prevents full utilization of unpaired data, leading to deficiencies in identity consistency and physical effect generation. In summary, 
\begin{itemize}
    \item We propose a diffusion-based solution for object removal/insertion with physical and geometric consistency in physical effects including shadows and reflections.
    
    \item We introduce a progressive training pipeline, where the proposed CycleFlow technique enables unpaired post-training, minimizing reliance on paired data.
    
    \item We further develop a novel no-reference metric called CFD for object removal quality through hallucination detection and context coherence assessment.
    
\end{itemize}

\begin{figure*}[t]
    \centering
    \includegraphics[width=\linewidth]{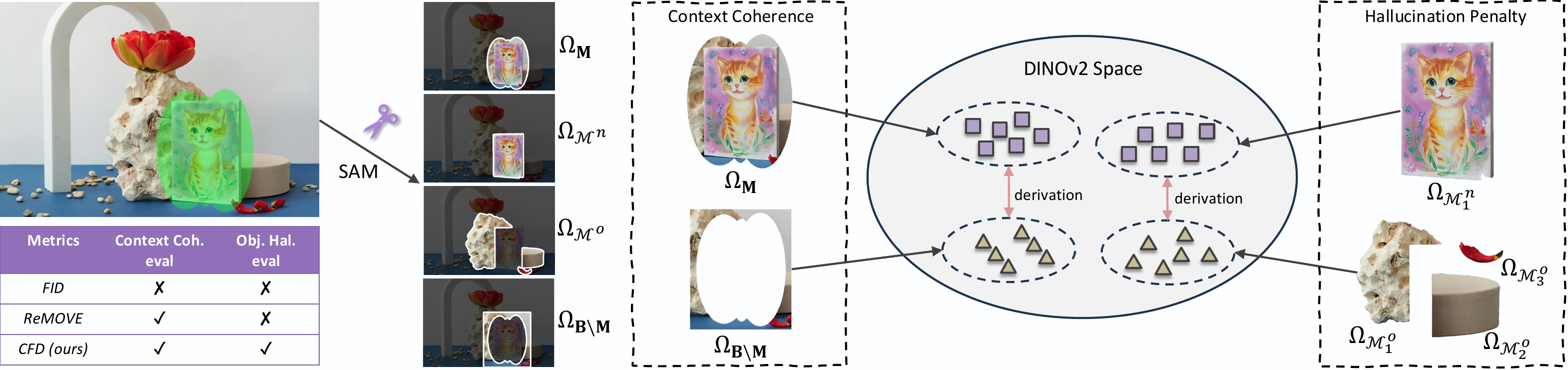}
    \vspace{-2mm}
    \caption{Illustration of the proposed CFD metric for evaluating object removal quality. Left: We apply SAM to segment the inpainted image into object masks and classify them into nested ($\Omega_{\mathcal{M}^{n}}$) and overlapping ($\Omega_{\mathcal{M}^{o}}$) masks. Middle: The context coherence term measures the feature deviation between the inpainted region ($\Omega_{\mathbf{M}}$) and its surrounding background ($\Omega_{\mathbf{B} \setminus \mathbf{M}}$) in the DINOv2 feature space. Right: The hallucination penalty is computed by comparing deep features of detected nested objects ($\Omega_{\mathcal{M}^{n}}$) with their adjacent overlapping masks ($\Omega_{\mathcal{M}^{o}}$) to assess whether unwanted object-like structures have emerged.}
    \label{fig:cfd-diagram}
    \vspace{-3mm}
\end{figure*}

\section{Related Works}

\subsection{Image Inpainting}
The task of image inpainting, specifically filling in missing pixels within masked regions, has been extensively studied in the literature. End-to-end learning approaches~\cite{li2022mat,suvorov2022resolution,yu2022unbiased}, which aim for pixel-wise fidelity, produce blurry or repetitive patterns when dealing with large masks or complex backgrounds. Methods leveraging pre-trained generative models, such as GANs~\cite{yu2022high,zheng2022image,yildirim2023diverse} and diffusion models~\cite{avrahami2023blended,wang2024magic}, as priors to generate realistic content for missing regions. Recently, based on text-to-image models~\cite{rombach2022high} have enabled controllable inpainting~\cite{powerpaint,FLUX-Inpainting,objectdrop}, allowing guided synthesis within masked regions.

\noindent\textbf{Key Differences from Conventional Inpainting.}
Our approach departs from traditional inpainting paradigms in two fundamental ways:

\begin{itemize}
    \item Standard inpainting reconstructs masked images to match the original, while we explicitly model object removal and insertion as distinct yet interdependent processes.
    \item Traditional inpainting fills hole, while our approach may adjust surrounding content for seamless integration.
\end{itemize}

\subsection{Realistic Object Removal}

Realistic object removal aims to eliminate foreground semantics while ensuring seamless background blending and preventing object hallucination. Existing methods fall into two categories: text-driven and mask-guided. 
Text-driven approaches~\cite{yu2024promptfix,yildirim2023inst,DBLP:conf/iclr/FuHDWYG24} specify objects for removal via instructions but are constrained by text embedding performance~\cite{DBLP:conf/cvpr/Marcos-ManchonA24,DBLP:conf/eccv/YangDJMWSJ24}, particularly in handling multiple objects and attribute understanding. Mask-guided methods~\cite{powerpaint, objectdrop, ekin2024clipaway, freecompose} provide more precise control. 
Recent advances, such as MagicEraser~\cite{magiceraser}, generate removal data by shifting objects within an image, while SmartEraser~\cite{jiang2025smarteraser} synthesizes a million-sample dataset using alpha blending. RORem~\cite{li2025rorem} leverages synthetic datasets like MULAN~\cite{tudosiu2024mulan} for training, though synthetic data limit a model's ability to replicate realistic object effects.

We introduce the CFD score, a reference-free metric designed exclusively for object removal, to evaluate object hallucination and context coherence. This enables a more effective assessment of object removal techniques.

\subsection{Generative Object Insertion}
Object insertion aims to seamlessly integrate new objects into existing scenes. Early methods focused on harmonization and blending~\cite{azadi2020compositional,tsai2017deep,wu2019gp} but struggled with complex physical effects like shadows and reflections. Recent approaches leverage real-world datasets~\cite{objectdrop}, synthetic blender-generated data~\cite{DBLP:journals/corr/abs-2412-02635}, or test-time tuning~\cite{freecompose} to improve object--background interactions, yet they remain limited in geometric alignment modeling.

Diffusion models offer a promising alternative, often using object-specific embeddings from CLIP~\cite{objectstitch,paintbyexample} or DINOv2~\cite{anydoor,imprint} to preserve identity, attributes, and texture. Unlike these existing works, our approach builds on FLUX without additional feature extractors.

Concurrent methods~\cite{objectmate,diptych} tackle generative object insertion along the direction of text-driven subject generation. ObjectMate~\cite{objectmate} constructs millions of paired samples covering multiple reference subjects. In contrast, we focus on \emph{single-subject} image-driven insertion, ensuring subject alignment and effect integration using only 3K real-world paired samples as training data, followed by CycleFlow unpaired post-training. This approach significantly alleviates the requirement for large paired datasets.

\section{Preliminaries}

Flow Matching (FM)~\cite{DBLP:conf/iclr/LipmanCBNL23} is a generative modeling framework that learns a velocity field \( u_t(\mathbf{z}_t) \) to map a source distribution \( p_0 \) to a target distribution \( p_1 \) through a time-dependent flow. 
The goal of FM is to train a neural network $\theta$ to make prediction \( u_t^\theta(\mathbf{z}_t) \) to approximate the velocity field \( u_t(\mathbf{z}_t) \). This is achieved by minimizing the \textit{Flow Matching Loss}, defined as:
\begin{equation}
    \mathcal{L}_{\text{FM}}(\theta) = \mathbb{E}_{t, \mathbf{z}_t \sim p_t} \left[ \| u_t^\theta(\mathbf{z}_t) - u_t(\mathbf{z}_t) \|^2 \right],
\end{equation}
where \( \mathbf{z}_t \sim p_t \) and \( t \sim \mathcal{U}[0, 1] \). Directly optimizing \( \mathcal{L}_{\text{FM}}(\theta) \) is intractable due to the complexity of estimating the ground truth velocity field \( u_t(\mathbf{z}_t) \) for arbitrary \( p_t \).

To simplify optimization, the Conditional Flow Matching (CFM) framework introduces conditional distributions \( p_{t|1}(\mathbf{z} | \mathbf{z}_1)= \mathcal{N}\left(\mathbf{z} | t\mathbf{z}_1, (1-t)^2I\right) \), which focuses on paths conditioned on target samples \( \mathbf{z}_1=Z_1\). The velocity field under this conditional setting is analytically given by:
\begin{equation}
    u_t(\mathbf{z} | \mathbf{z}_1) = \frac{\mathbf{z}_1 - \mathbf{z}}{1 - t}.
\end{equation}
The conditional probability path \( p_{t|1}(\mathbf{z} | \mathbf{z}_1) \) follows a linear interpolation:
\begin{equation}
    \mathbf{z}_{t} = t\mathbf{z}_1 + (1-t)\mathbf{z}_0 \quad \mathbf{z}_t=Z_t \sim p_{t|1}(\cdot | \mathbf{z}_1)
\end{equation}
where \( \mathbf{z}_0=Z_0 \sim p_0 \) and \( \mathbf{z}_1=Z_1 \sim p_1 \). Using this formulation, the \textit{Conditional Flow Matching Loss} is defined as:
\begin{equation}\label{eqn:cfm_origin}
    \mathcal{L}_{\text{CFM}}(\theta) = \mathbb{E}_{t, \mathbf{z}_0 \sim p_0, \mathbf{z}_1 \sim p_1} \left[ \| u_t^\theta(\mathbf{z}_t) - u_t(\mathbf{z}_t | \mathbf{z}_1) \|^2 \right].
\end{equation}

This loss avoids the need to estimate \( u_t(\mathbf{z}_t) \) directly by leveraging the known form of \( u_t(\mathbf{z} | \mathbf{z}_1) \).

\section{Methodology}
We frame image inpainting as a dual-path, object-oriented process that consists of two key directions: \textbf{object removal} and \textbf{object insertion}. Given an image $\mathbf{I} \in \mathbb{R}^{H \times W \times 3}$ and a binary mask $\mathbf{M} \in \{0,1\}^{H \times W}$ denoting the edited region (where $\mathbf{M}_{ij} = 1$ indicates masked pixels), our model operates on the masked input $\mathbf{X} = \mathbf{I} \odot (1 - \mathbf{M})$ to facilitate targeted modifications.
The object removal pathway suppresses semantic traces within $\mathbf{M}$, ensuring smooth boundary transitions while preventing unintended artifacts or hallucinations. Meanwhile, the object insertion pathway integrates a new object $\mathbf{O} \in \mathbb{R}^{H' \times W' \times 3}$ ($H' < H, W' < W$), maintaining global coherence and context-aware realism.

\subsection{The OmniPaint Framework}

OmniPaint builds upon FLUX-1.dev~\cite{flux2024}, a diffusion-based architecture featuring a Multi-Modal Diffusion Transformer (MM-DiT)~\cite{sd3} backbone. While preserving \texttt{FLUX}'s strong text-to-image priors, we introduce the image-conditioning mechanisms used in ~\cite{tan2024ominicontrol} tailored for object-aware editing.

\noindent\textbf{Masked Image Conditioning.}
The model refines Gaussian noise $\mathbf{z}_0 =Z_0\sim p_0$ towards $\mathbf{z}_1$, using the masked image $\mathbf{X}$ as a denoising guide for object removal and insertion. 
We leverage the existing \texttt{FLUX} networks, including its VAE encoder and $2\times2$ patchify layer, to map $\mathbf{X}$ into a shared feature space, yielding the conditioned token sequence $\mathbf{z}_c^{\mathcal{X}}$.

\noindent\textbf{Reference Object Conditioning.}
For object insertion, the model conditions on both the masked image and a reference object image $\mathbf{O}$. 
To preserve object identity while minimizing background interference, we preprocess $\mathbf{O}$ with Carvekit~\cite{Selin2023} for background removal before resizing it to match $\mathbf{X}$'s spatial dimensions. The reference object undergoes the same latent encoding and patchification as the masked image, producing a corresponding latent sequence $\mathbf{z}_c^{\mathcal{O}}$. The final condition token is obtained by concatenating both sequences along the token dimension: $\mathbf{z}_c = [\mathbf{z}_c^{\mathcal{X}}; \mathbf{z}_c^{\mathcal{O}}]$.

\noindent\textbf{Prompt-Free Adaptive Control.}
Given the highly image-conditioned nature of our task, textual prompts may introduce ambiguity. To mitigate this, we adopt a prompt-free adaptive control mechanism, replacing text embeddings with learnable task-specific parameters. Specifically, we introduce two trainable vectors:
\begin{equation}
\tau_{\text{removal}}, \tau_{\text{insertion}} \sim \mathcal{N}(0, I),
\end{equation}
initialized from the embedding of an empty string and optimized separately for each task. Inference switches between removal and insertion via embedding selection.

To facilitate computational efficiency, we freeze the \texttt{FLUX} backbone and perform Parameter-Efficient Fine-Tuning (PEFT), optimizing two LoRA~\cite{DBLP:conf/iclr/HuSWALWWC22} parameter sets, $\theta$ and $\phi$, for object removal and insertion, respectively.

\subsection{Data Collection and Mask Augmentation}
We collect a dataset of 3,300 real-world paired samples captured across diverse indoor and outdoor environments, encompassing various physical effects such as shadows, specular reflections, optical distortions, and occlusions (see Appendix for examples). Each triplet $\langle\mathbf{I}, \mathbf{I}_{\text{removed}}, \mathbf{M}\rangle$ is meticulously annotated to ensure high quality.

To enhance model robustness against diverse mask variations, we apply distinct augmentation strategies for object removal and insertion. For removal, we introduce segmentation noise via morphological transformations, randomly applying dilation or erosion with configurable parameters. Imprecise masks are simulated by perturbing boundaries and adding or removing geometric shapes (e.g., circles, rectangles). Augmented examples 
and the effectiveness analysis are provided in the Appendix.
For object insertion, since explicit object detection is not required, we simplify mask augmentation by expanding segmentation masks to their bounding boxes or convex hulls, ensuring adaptability to various reference object formats. Reference object image augmentation follows prior work~\cite{imprint}.

\subsection{Training Pipeline}

In our experiments, we observe that the current training data are insufficient to maintain reference identity for object insertion, as in Fig.~\ref{fig:ablation_hyperparameters}(b) and Table~A in the Appendix.
Bootstrapping paired data via trained models, akin to ObjectDrop~\cite{objectdrop}, is a straightforward solution but requires a reliable filtering mechanism, which remains an open challenge.

Fortunately, object insertion and object removal are mathematically complementary inverse problems (i.e., each can be viewed as inverting the other). Inspired by cycle-consistency approaches~\cite{DBLP:conf/iccv/ZhuPIE17,DBLP:conf/iccv/WuT23}, we propose utilizing \emph{unpaired} data rather than relying on paired augmentations. In particular, we utilize large-scale object segmentation datasets, which lack explicit removal pairs, to enhance object insertion. This section presents our three-phase training pipeline:
\emph{(1) inpainting pretext training}, 
\emph{(2) paired warmup}, 
and 
\emph{(3) CycleFlow unpaired post-training}.

\subsubsection{Inpainting Pretext Training}
To endow our model with basic inpainting abilities, we first fine-tune it on a pretext inpainting task, initializing $\theta$ and $\phi$ for later stages.
Using a mask generator~\cite{suvorov2022resolution}, we apply random masks to LAION dataset~\cite{DBLP:conf/nips/SchuhmannBVGWCC22} and train the model to reconstruct missing regions by minimizing a CFM loss,
\begin{equation}
\label{eq:inpainting-loss}
\begin{aligned}
    \mathcal{L}_{\text{pretext}}&(\theta, \phi | \mathbf{z}_t, \mathbf{z}_c^{\mathcal{X}}) = \\
    & \mathbb{E}_{t, \mathbf{z}_0, \mathbf{z}_1} \left[ \| u_t^{\theta, \phi}(\mathbf{z}_t, \mathbf{z}_c^{\mathcal{X}}) - u_t(\mathbf{z}_t | \mathbf{z}_1) \|^2 \right],
\end{aligned}
\end{equation}
where \( \mathbf{z}_1 = Z_1 \sim p_1(\mathbf{I}) \), which enforces the model to complete the masked region so that the entire image approximates \( \mathbf{I} \).
We show in the appendix that pretext training benefits object editing performance.

\subsubsection{Paired Warmup}
Next, we leverage our 3,000 paired samples for real-world object insertion and removal training. In the Paired Warmup stage, $\theta$ and $\phi$ are trained separately, enabling effect-aware object removal (e.g., removing reflections and shadows) and insertion with effect integration.

For insertion, \(\mathbf{z}_1\) is drawn from \(Z_1 \sim p_1(\mathbf{I})\), where \(I\) means images retaining the foreground object. We optimize the following objective by modifying Equation~\ref{eqn:cfm_origin}:
\begin{equation}\label{eqn:cfm_cond_insert}
\begin{aligned}
    \mathcal{L}_{\text{warmup}}&(\theta | \mathbf{z}_t, \mathbf{z}_c, \tau) = \\
    & \mathbb{E}_{t, \mathbf{z}_0, \mathbf{z}_1} \left[ \| u_t^{\theta}(\mathbf{z}_t, \mathbf{z}_c, \tau) - u_t(\mathbf{z}_t | \mathbf{z}_1) \|^2 \right],
\end{aligned}
\end{equation}
where \(\mathbf{z}_c = [\mathbf{z}_c^{\mathcal{X}}; \mathbf{z}_c^{\mathcal{O}}]\) represents the conditioning token sequence, concatenating masked image and object identity features, and $\tau$ denotes the corresponding task-specific embedding.

For removal, \(\mathbf{z}_1\) is sampled from \(Z_1 \sim p_1(\mathbf{I}_{\text{removed}})\), where \(\mathbf{I}_{\text{removed}}\) means images with the foreground object physically removed. Given conditioning on \(\mathbf{z}_c^{\mathcal{X}}\), the optimization objective becomes:
\begin{equation}\label{eqn:cfm_cond_remove}
\begin{aligned}
    \mathcal{L}_{\text{warmup}}&(\phi | \mathbf{z}_t, \mathbf{z}_c^{\mathcal{X}}, \tau) = \\
    & \mathbb{E}_{t, \mathbf{z}_0, \mathbf{z}_1} \left[ \| u_t^{\phi}(\mathbf{z}_t, \mathbf{z}_c^{\mathcal{X}}, \tau) - u_t(\mathbf{z}_t | \mathbf{z}_1) \|^2 \right].
\end{aligned}
\end{equation}

In practice, we assume a linear interpolation path for computational efficiency~\cite{DBLP:conf/iclr/LiuG023}, setting \( u_t(\mathbf{z}_t | \mathbf{z}_1) = (\mathbf{z}_1 - \mathbf{z}_0) \) in both objectives.
This warmup stage enhances object removal, effectively handling reflections and shadows (Fig.\ref{fig:qual_remove}). However, with only 3,000 paired samples, it struggles to maintain reference identity in object insertion (Fig.\ref{fig:ablation_hyperparameters}(b)).

\begin{figure}
    \centering
    \includegraphics[width=0.8\linewidth]{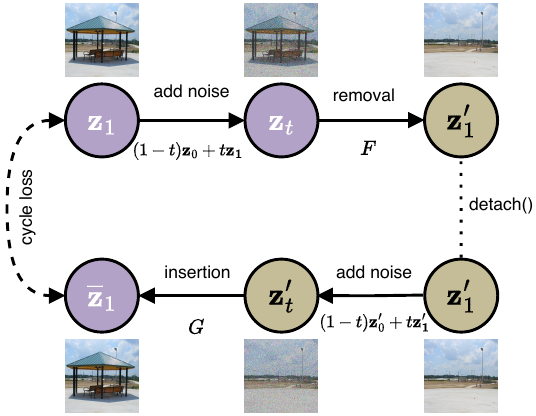}
    \vspace{-2mm}
    \caption{
    Illustration of CycleFlow. The mapping $F$ removes the object, predicting an estimated target $\mathbf{z}_1'$, while $G$ reinserts the object, generating estimated target $\overline{\mathbf{z}}_1$. Cycle consistency is enforced by ensuring $G$ reconstructs the original latent $\mathbf{z}_1$ from the effect removal output. Dashed arrows indicate the cycle loss supervision.}
    \label{fig:cycleflow}
    \vspace{-3mm}
\end{figure}

\subsubsection{CycleFlow Unpaired Post-Training}

To enhance training for object insertion, we leverage large-scale object segmentation datasets, including COCO-Stuff~\cite{COCO-stuff} and HQSeg~\cite{sam_hq}, as unpaired data sources. These datasets provide foreground object masks, enabling us to easily construct the model’s conditioning inputs $\mathbf{X}$ and $\mathbf{O}$. 

We continue tuning $\theta$ using the same objective as in Equation~\ref{eqn:cfm_cond_insert} on this larger dataset, improving identity preservation, as shown in Fig.~\ref{fig:ablation_hyperparameters}(b). The case where $\gamma=0$ corresponds to training solely with Equation~\ref{eqn:cfm_cond_insert}. However, these segmentation datasets lack annotations for object effects, such as shadows and reflections, meaning that the masked image input $\mathbf{X}$ still retains these effects. This suppresses the model’s ability to synthesize realistic object effects, making insertions appear more like copy-paste operations of the reference object, as observed in the $\gamma=0$ case of Fig.~\ref{fig:ablation_hyperparameters}(b).

To overcome this limitation, we use our well-trained removal parameters $\phi$, which even at $\text{NFE}=1$ remove object effects (See Fig.~\ref{fig:ablation_hyperparameters}(a)). Leveraging $\phi$ as a preprocessing step enables insertion training on latents with effects removed.

Thus, we introduce the \emph{CycleFlow} mechanism, comprising two mappings: $F$ (removal direction) and $G$ (insertion direction). These mappings predict the velocity field at $\mathbf{z}_t$, estimating their target samples $\mathbf{z}_1 =Z_1, Z_1 \sim p_1$:
\begin{align}
    F: & \mathbf{z}_1^\prime \leftarrow \mathbf{z}_t - u^\phi_t(\mathbf{z}_t, \mathbf{z}_c^{\mathcal{X}},\tau_\text{removal}) \cdot t, \\
    G: & \overline{\mathbf{z}}_1 \leftarrow \mathbf{z}^\prime_t - u^\theta_t(\mathbf{z}^\prime_t, \mathbf{z}_c,\tau_\text{insertion}) \cdot t,
\end{align}
where $\mathbf{z}_1^\prime$ and $\overline{\mathbf{z}}_1$ denote the estimated target samples for removal and insertion, respectively. Here, we also rely on the \( u_t(\mathbf{z}_t | \mathbf{z}_1) = (\mathbf{z}_1 - \mathbf{z}_0) \) linear interpolation setting~\cite{DBLP:conf/iclr/LiuG023}.

As illustrated in Fig.~\ref{fig:cycleflow}, we design a \emph{Remove-Insert} cycle, ensuring that reinserting a removed object approximately restores its original latent representation.
\begin{equation}
    \mathbf{z}_1 \rightarrow \mathbf{z}_t \rightarrow F(\mathbf{z_t}) \rightarrow \mathbf{z}^\prime_t \rightarrow G(\mathbf{z}^\prime_t) \approx \mathbf{z}_1.
\end{equation}
To enforce this cycle consistency, we define a \emph{Cycle Loss}:
\begin{equation}
\mathcal{L}_{\text{cycle}} (\theta)
= 
\mathbb{E}_{t,\,\mathbf{z}_t} 
\Bigl[
  \bigl\|\,G_{\theta}\!\bigl(\lfloor F(\mathbf{z}_t)\rfloor \bigr) - \mathbf{z}_1 \bigr\|^{2}
\Bigr],
\end{equation}
where \(\lfloor \cdot \rfloor\) denotes a gradient truncation operator, treating its output as a constant during backprop to fix parameters $\phi$. 
During CycleFlow post-training, we optimize an overall loss:
$\mathcal{L}_\text{warmup}(\theta |\mathbf{z}_t, \mathbf{z}_c, \tau_\text{insertion}) + \gamma \mathcal{L}_\text{cycle}$
on unpaired training data, where $\gamma$ controls the strength of cycle consistency (analyzed in Sec.~\ref{sec:ablation}).

Empirically, this work focuses solely on CycleFlow for object insertion, as warmup-alone suffices for removal. 

\subsection{Context-Aware Feature Deviation (CFD) Score}
\label{sec:cfd_score}

We introduce the Context-Aware Feature Deviation (CFD) score to quantitatively assess object removal performance. As illustrated in Fig.~\ref{fig:cfd-diagram}, CFD comprises two components: a hallucination penalty that detects and penalizes unwanted object-like structures emerging in the removed region, and a context coherence term that evaluates how well the inpainted region blends with the surrounding background.

\noindent\textbf{Hallucination Penalty.} 
Given an object mask $\mathbf{M}$, let $\Omega_{\mathbf{M}} = \{(i, j)\,\mid\,\mathbf{M}_{ij} = 1\}$ denote the pixels of removed region. Define $\mathbf{B} = \texttt{bbox}(\mathbf{M})$ as its bounding box. After removal, we aim to identify whether the synthesized content introduces spurious object-like structures.

We apply the off-the-shelf SAM-ViT-H~\cite{sam1} model to segment the image into masks $\{\mathcal{M}_k\}_{k=1}^{K}$. Focusing on masks near $\mathbf{M}$, we categorize them as:
\begin{itemize}
    \item \emph{Nested masks}, $\mathcal{M}^{\mathrm{n}} = \{\mathcal{M}_k^n \mid \Omega_{\mathcal{M}_k^n} \subseteq \Omega_{\mathbf{M}}\}$, entirely contained within the removed region.
    \item \emph{Overlapping masks}, $\mathcal{M}^{\mathrm{o}} = \{\mathcal{M}_k^o \mid \Omega_{\mathcal{M}_k^o} \cap \Omega_{\mathbf{M}} \neq \varnothing, \Omega_{\mathcal{M}_k^o} \not\subseteq \Omega_{\mathbf{M}}\}$, partially overlapping $\Omega_{\mathbf{M}}$ but extending beyond.
\end{itemize}

A naive hallucination penalty would simply count nested masks, but some may arise from segmentation noise. Instead, we leverage \emph{deep feature similarity} to assess whether a mask plausibly integrates into its context. To refine segmentation, we merge overlapping masks adjacent to any $\mathcal{M}_i^n \in \mathcal{M}^{\mathrm{n}}$:
\begin{equation}
\mathcal{M}^{\mathrm{paired}} = \Bigl\{(\mathcal{M}_i^n, \overline{\mathcal{M}}_i)
\,\mid\, 
\overline{\mathcal{M}}_i = 
\!\!\bigcup_{ \mathrm{adj}(\mathcal{M}_j^o, \mathcal{M}_i^n)} \mathcal{M}_j^o
\Bigr\},
\end{equation}
where $\mathcal{M}_j^o \in \mathcal{M}^{\mathrm{o}}$ denotes a overlapping mask, and $\mathrm{adj}(\mathcal{M}_j^o, \mathcal{M}_i^n) = 1$ if the masks share a boundary pixel or their one-pixel dilation overlaps.

The hallucination penalty is then defined as:
\begin{equation}
d_{\text{hallucination}} 
= 
\sum_{(\mathcal{M}_i^n, \overline{\mathcal{M}}_i) \in \mathcal{M}^{\mathrm{paired}}}
       \omega_i \cdot \Bigl(1 - \mathbf{f}(\Omega_{\mathcal{M}_i^n})^\top \mathbf{f}(\Omega_{\overline{\mathcal{M}}_i})\Bigr),
\label{eq:cfd_hallu_final}
\end{equation}
where $\omega_i = \frac{|\Omega_{\mathcal{M}_i^n}|}{\sum_{\mathcal{M}^{\mathrm{n}}}|\Omega_{\mathcal{M}^{\mathrm{n}}}|}$ weights the contribution of each nested mask. Feature embeddings $\mathbf{f}(\Omega)$ are extracted from the pre-trained vision model DINOv2~\cite{dinov2}.

\begin{figure}[t]
    \centering
    \includegraphics[width=\linewidth]{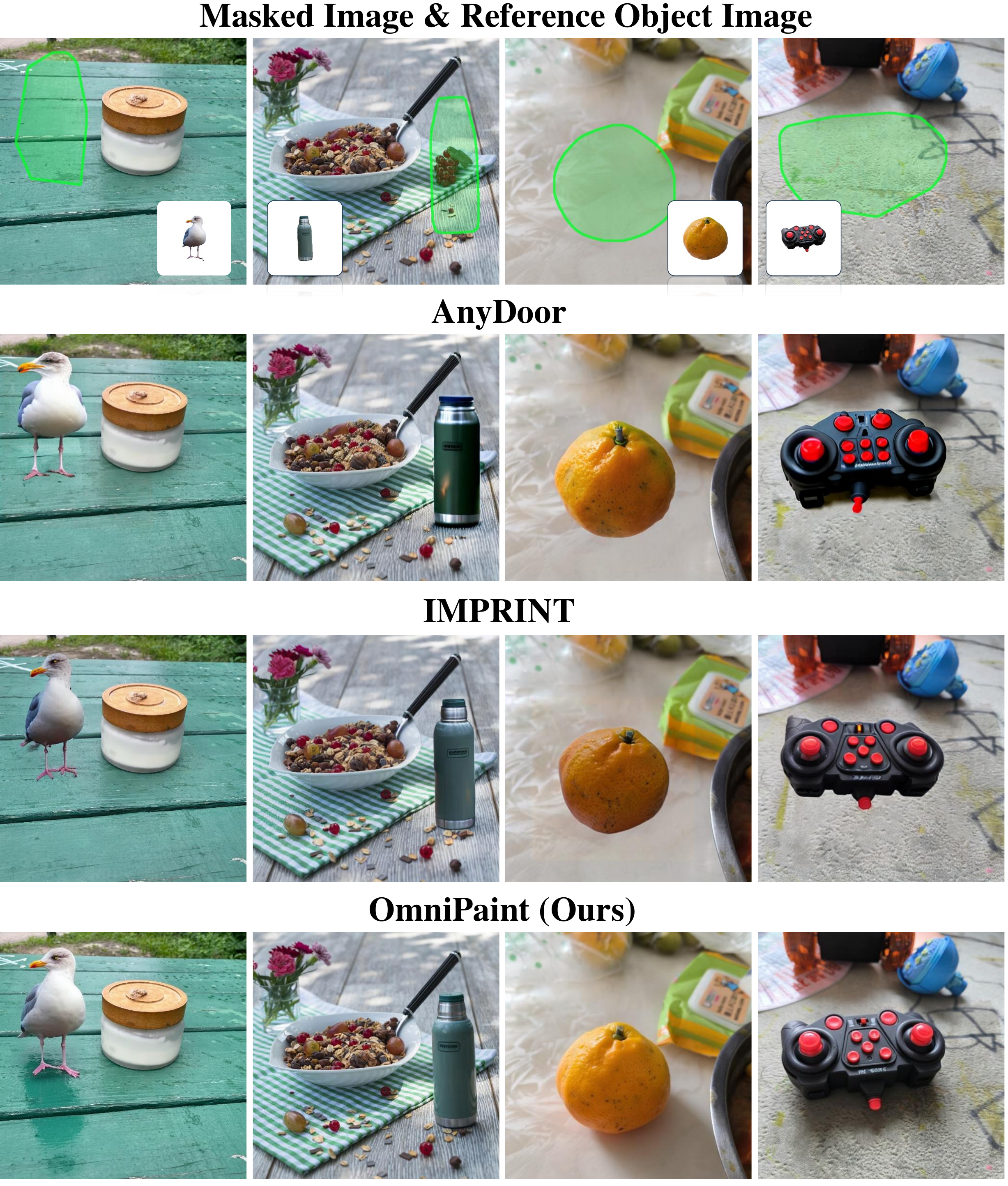}
    \vspace{-6mm}
    \caption{Qualitative comparison on object insertion. Given masked images and reference object images (top row), we compare results from AnyDoor~\cite{anydoor}, IMPRINT~\cite{imprint}, and OmniPaint.}
    \label{fig:qual_insert}
    \vspace{-4mm}
\end{figure}

\begin{figure*}[t]
    \centering
    \includegraphics[width=\linewidth]{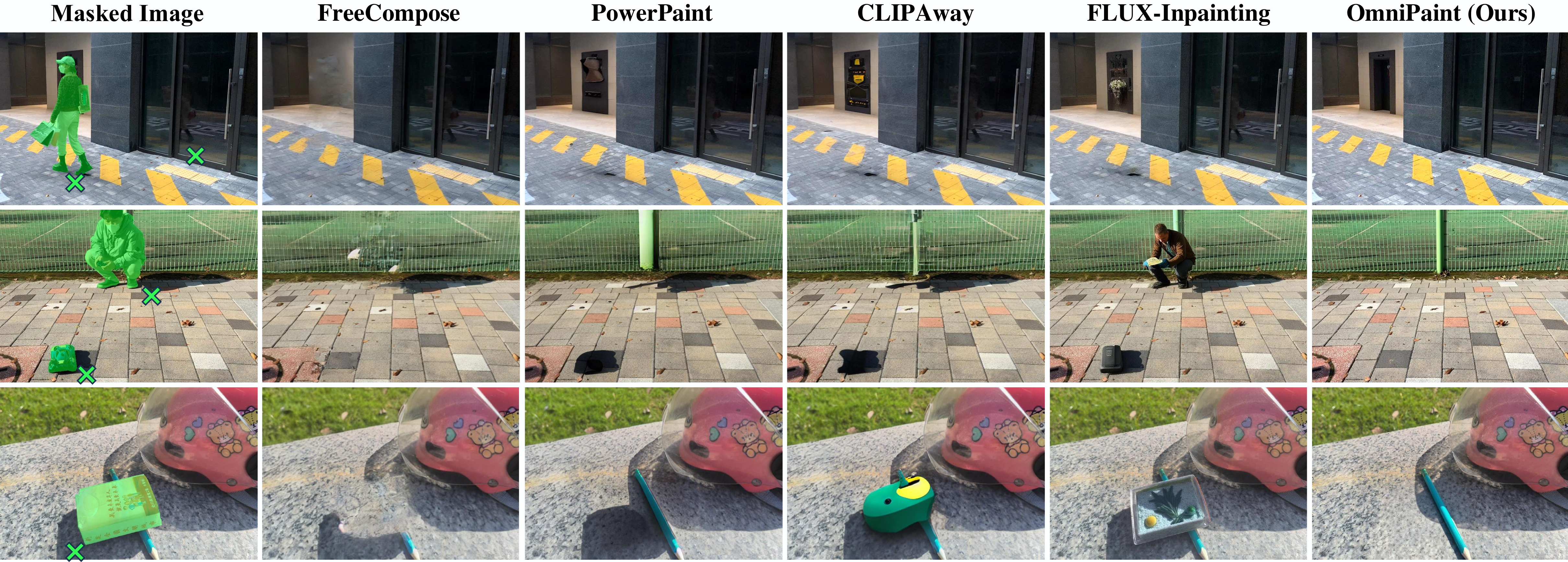}
    \vspace{-5mm}
    \caption{Qualitative comparison of object removal in challenging scenarios. \textbf{Top:} Simultaneous removal of objects and glass reflections. \textbf{Middle:} Shadow-free removal under real-world lighting. \textbf{Bottom:} Occlusion-robust inpainting, reconstructing background objects without distortion. The compared methods include  FreeCompose~\cite{freecompose}, PowerPaint~\cite{powerpaint}, CLIPAway~\cite{ekin2024clipaway}, and FLUX-Inpainting~\cite{FLUX-Inpainting}.}
    \label{fig:qual_remove}
    \vspace{-4mm}
\end{figure*}

\noindent\textbf{Context Coherence.} 
Even when $d_{\text{hallucination}} = 0$ (i.e., no nested objects are detected), the inpainted content may still not align with the surrounding background. To quantify this structural consistency, we compute feature deviation:
\begin{equation}
d_{\text{context}} 
= 
1 - \mathbf{f}(\Omega_{\mathbf{M}})^\top \mathbf{f}(\Omega_{\mathbf{B} \setminus \mathbf{M}}),
\end{equation}
where $\mathbf{B} \setminus \mathbf{M}$ denotes the bounding box excluding the masked region. 

\noindent\textbf{Final CFD Metric.}
The final CFD score is computed as:
\begin{equation}
\text{CFD} = d_{\text{context}} + d_{\text{hallucination}}.
\end{equation}
A lower CFD signifies better removal quality—minimal hallucination and seamless contextual blending.

\section{Experiments}
\label{sec:experiments}

\subsection{CFD Analysis}
We perform qualitative analyses to determine whether our CFD score effectively captures both contextual coherence and hallucination artifacts, thereby offering a more reliable evaluation of object removal quality compared to existing metrics such as ReMOVE~\cite{REMOVE}.
As illustrated in Fig.~\ref{fig:cfd_analysis}, FLUX-Inpainting~\cite{FLUX-Inpainting} generates conspicuous hallucinations—phantom objects like ships, human figures, or floating canisters—yet still attains high ReMOVE scores. In contrast, CFD effectively penalizes these hallucinations by using SAM to segment the inpainted region and by examining feature-level discrepancies within nested and overlapping masks. Similarly, while LaMa~\cite{suvorov2022resolution} interpolates background textures in the masked area, its limited generative prior often leads to ghostly artifacts due to insufficient object effect detection. Conversely, our OmniPaint demonstrates superior removal fidelity by completely eliminating target objects without introducing unwanted artifacts, as reflected by its significantly lower CFD scores. 

By concurrently quantifying both the emergence of unwanted objects and contextual alignment, CFD aligns closely with human visual perception. These findings substantiate CFD as a robust evaluation metric that helps ensure that object removal not only achieves seamless blending but also minimizes erroneous content hallucination.

\begin{table}[t]
    \centering
    \caption{Quantitative results on our 300-sample removal test set.}
    \vspace{-2mm}
    \label{tab:quan_compare_ourtestset}
    \resizebox{\linewidth}{!}{
    \begin{tabular}{lccccccc}
        \toprule
        \textbf{Method} & \textbf{FID} $\downarrow$ & \textbf{CMMD} $\downarrow$ & \textbf{CFD} $\downarrow$ & \textbf{ReMOVE} $\uparrow$ & \textbf{PSNR} $\uparrow$ & \textbf{SSIM} $\uparrow$ & \textbf{LPIPS} $\downarrow$ \\
        \midrule
        LaMa         & 105.10 & 0.3729 & \underline{0.3531} & 0.7311 & 20.8632 & \textbf{0.8278} & 0.1353 \\
        MAT            & 147.37 & 0.6646 & 0.5104 & 0.6162 & 18.2229 & 0.7845 & 0.1900 \\
        SD-Inpainting   & 153.13 & 0.3997 & 0.4874 & 0.6234 & 18.8760 & 0.6932 & 0.1830 \\
        FLUX-Inpainting & 132.60 & 0.3257 & 0.4609 & 0.6765 & 20.8560 & 0.8002 & 0.1451 \\
        CLIPAway      & 115.72 & 0.2919 & 0.5242 & 0.7396 & 19.5259 & 0.7085 & 0.1641 \\
        PowerPaint    & 103.61 & 0.2182 & 0.4031 & 0.8013 & 19.4559 & 0.7102 & 0.1428 \\
        FreeCompose   & \underline{88.77}  & \underline{0.1790} & 0.3743 & \textbf{0.8654} & \underline{21.2729} & 0.7320 & \underline{0.1182} \\
        \textbf{OmniPaint (Ours)} & \textbf{51.66}  & \textbf{0.0473} & \textbf{0.2619} & \underline{0.8610} & \textbf{23.0797} & \underline{0.8135} & \textbf{0.0738} \\
        \bottomrule
    \end{tabular}}
    \vspace{-4mm}
\end{table}

\subsection{Experimental Settings}

For removal, we compare against end-to-end inpainting models MAT~\cite{li2022mat} and LaMa~\cite{suvorov2022resolution}, the diffusion-based SD-Inpaint~\cite{rombach2022high}, and FLUX-Inpainting~\cite{FLUX-Inpainting} to ensure a fair backbone comparison. Additionally, we include recent open-source object removal methods CLIPAway~\cite{ekin2024clipaway}, PowerPaint~\cite{powerpaint}, and FreeCompose~\cite{freecompose}. Experiments are conducted on two benchmarks: a test set of 300 real-world object removal cases we captured, resized to $512^2$ for testing, and the RORD~\cite{rord} dataset with 1,000 paired samples at their original $540 \times 960$ resolution, both providing ground truth from physically removed objects. We report PSNR, SSIM, perceptual similarity metrics (FID~\cite{fid}, CMMD~\cite{DBLP:conf/cvpr/JayasumanaRVGCK24}, LPIPS~\cite{DBLP:conf/cvpr/ZhangIESW18}), and object removal-specific metrics, including ReMOVE~\cite{REMOVE} and our CFD score.

For object insertion, we compare against Paint-by-Example (PbE)~\cite{paintbyexample}, ObjectStitch~\cite{objectstitch}, FreeCompose~\cite{freecompose}, AnyDoor~\cite{anydoor}, and IMPRINT~\cite{imprint}. Since ObjectStitch and IMPRINT do not have public implementations, we obtain official code, checkpoints, and test sets from the authors. Our insertion benchmark consists of 565 samples with $512^2$ resolutions, combining the IMPRINT test set with real-world cases we captured. Each sample includes a background image, a reference object image, and a binary mask. Reference images are preprocessed using CarveKit~\cite{Selin2023} for background removal.
To evaluate identity consistency, we measure feature similarity between the inserted object and its reference counterpart using CUTE~\cite{DBLP:conf/nips/KotarTYY023}, CLIP-I~\cite{DBLP:conf/icml/RadfordKHRGASAM21}, DINOv2~\cite{dinov2}, and DreamSim~\cite{DBLP:conf/nips/FuTSC0DI23}, with the latter being more aligned with human perception. Beyond local identity preservation, we assess overall image quality using non-reference metrics: MUSIQ~\cite{DBLP:conf/iccv/KeWWMY21} and MANIQA~\cite{DBLP:conf/cvpr/YangWSLGCWY22}.

\begin{table}[t]
    \centering
    \caption{Quantitative results on the 1000-sample RORD test set.}
    \vspace{-2mm}
    \label{tab:quan_compare_rord}
    \resizebox{\linewidth}{!}{
    \begin{tabular}{lccccccc}
        \toprule
        \textbf{Method} & \textbf{FID} $\downarrow$ & \textbf{CMMD} $\downarrow$ & \textbf{CFD} $\downarrow$ & \textbf{ReMOVE} $\uparrow$ & \textbf{PSNR} $\uparrow$ & \textbf{SSIM} $\uparrow$ & \textbf{LPIPS} $\downarrow$ \\
        \midrule
        LaMa          & 49.20  & 0.4897  & 0.4660  & 0.8321  & 19.2941  & 0.5571  & 0.1075 \\
        MAT           & 86.33  & 0.8689  & 0.7723  & 0.7070  & 20.3080  & \underline{0.7815}  & 0.1429 \\
        SD-Inpainting    & 75.31  & 0.4733  & 0.6648  & 0.8227  & 19.8308  & 0.6233  & 0.1235 \\
        FLUX-Inpainting & 62.24  & \underline{0.3805}  & 0.6077  & 0.8461  & \underline{21.9159}  & 0.7769  & 0.0975 \\
        CLIPAway      & 49.07  & 0.4569  & 0.5442  & 0.8696  & 20.3077  & 0.6055  & 0.1132 \\
        PowerPaint    & \underline{42.65}  & 0.4599  & \underline{0.4128}  & 0.8933  & 20.1832  & 0.6066  & \underline{0.0968} \\
        FreeCompose   & 46.37  & 0.5125  & 0.5215  & \underline{0.9008}  & 20.5678  & 0.6152  & 0.1090 \\
        \textbf{OmniPaint (Ours)} & \textbf{19.17}  & \textbf{0.2239}  & \textbf{0.3682}  & \textbf{0.9053}  & \textbf{23.2334}  & \textbf{0.7867}  & \textbf{0.0424} \\
        \bottomrule
    \end{tabular}}
    \vspace{-3mm}
\end{table}

For fairness, we apply the same image-mask pairs across all baselines and use official implementations with their default hyperparameters, such as inference step counts. For OmniPaint, we employ the Euler Discrete Scheduler~\cite{sd3} during inference and set the number of inference steps to $28$ for primary quantitative and qualitative experiments. Additional implementation details are provided in the Appendix.

\subsection{Evaluation of Object Removal Performance}

We evaluate OmniPaint on realistic object removal, comparing against inpainting and object removal methods. As shown in Table~\ref{tab:quan_compare_ourtestset} and Table~\ref{tab:quan_compare_rord}, OmniPaint consistently outperforms prior approaches across all datasets, achieving the lowest FID~\cite{fid}, CMMD~\cite{DBLP:conf/cvpr/JayasumanaRVGCK24}, LPIPS~\cite{DBLP:conf/cvpr/ZhangIESW18}, and CFD while maintaining high PSNR, SSIM, and ReMOVE~\cite{REMOVE} scores. These results highlight its ability to remove objects while preserving structural and perceptual fidelity, effectively suppressing object hallucination.

Fig.~\ref{fig:qual_remove} provides a visual comparison in challenging real-world cases. In the first row, OmniPaint successfully removes both objects and their glass reflections, a failure by all baselines. The second row highlights OmniPaint’s ability to eliminate shadows under natural lighting, where other methods leave residual artifacts. The third row demonstrates robust inpainting in occlusion scenarios, ensuring seamless background reconstruction without distortion.

By effectively handling reflections, shadows, and occlusions, OmniPaint surpasses prior methods in generating coherent and realistic object removal results.

\begin{table}[t]
    \centering
    \caption{Quantitative comparison of object insertion methods.}
    \vspace{-2.2mm}
    \label{tab:quan_compare_insert}
    \resizebox{\linewidth}{!}{
    \begin{tabular}{l|cccc|cc}
        \toprule
        & \multicolumn{4}{|c|}{\textit{Object Identity Preservation}} & \multicolumn{2}{c}{\textit{Overall Image Quality}} \\
         & \textbf{CLIP-I} $\uparrow$ & \textbf{DINOv2} $\uparrow$ & \textbf{CUTE} $\uparrow$ & \textbf{DreamSim} $\downarrow$ & \textbf{MUSIQ} $\uparrow$ & \textbf{MANIQA} $\uparrow$ \\
        \midrule
        PbE         & 84.1265 & 50.0008 & 65.1053 & 0.3806 & \underline{70.26} & \underline{0.5088} \\
        ObjectStitch & 86.4506 & 59.6560 & 74.0478 & 0.3245 & 68.87 & 0.4755 \\
        FreeCompose & 88.1679 & 76.0085 & 82.8641 & 0.2134 & 66.67 & 0.4775 \\
        Anydoor      & 89.2610 & 76.9560 & 85.2566 & 0.2208 & 69.28 & 0.4593 \\
        IMPRINT     & \underline{90.6258} & \underline{76.8940} & \underline{86.1511} & \underline{0.1854} & 68.72 & 0.4711 \\
        \textbf{OmniPaint} & \textbf{92.2693} & \textbf{84.3738} & \textbf{90.2936} & \textbf{0.1557} & \textbf{70.59} & \textbf{0.5209} \\
        \bottomrule
    \end{tabular}
    }
    \vspace{-4mm}
\end{table}

\subsection{Evaluation of Object Insertion Performance}
We evaluate OmniPaint on object insertion, comparing it with advanced methods. As shown in Table~\ref{tab:quan_compare_insert}, OmniPaint achieves the highest scores across all object identity preservation metrics, including CLIP-I~\cite{DBLP:conf/icml/RadfordKHRGASAM21}, DINOv2~\cite{dinov2}, CUTE~\cite{DBLP:conf/nips/KotarTYY023}, and DreamSim~\cite{DBLP:conf/nips/FuTSC0DI23}, demonstrating superior alignment with the reference object. Additionally, it outperforms all baselines in overall image quality, as measured by MUSIQ~\cite{DBLP:conf/iccv/KeWWMY21} and MANIQA~\cite{DBLP:conf/cvpr/YangWSLGCWY22}, indicating better perceptual realism and seamless integration.

Fig.~\ref{fig:qual_insert} presents visual comparisons. Given a masked input and a reference object, OmniPaint generates inserted objects with more accurate shape, texture, and lighting consistency. In contrast, other methods struggle with identity distortion, incorrect shading, or noticeable blending artifacts. Notably, OmniPaint preserves fine details while ensuring the inserted object naturally aligns with scene geometry and illumination. By maintaining high-fidelity identity preservation and improving perceptual quality, OmniPaint sets a new standard for realistic object insertion.

\subsection{Hyperparameter Analysis}\label{sec:ablation}

\begin{figure}[t]
    \centering
    \includegraphics[width=\linewidth]{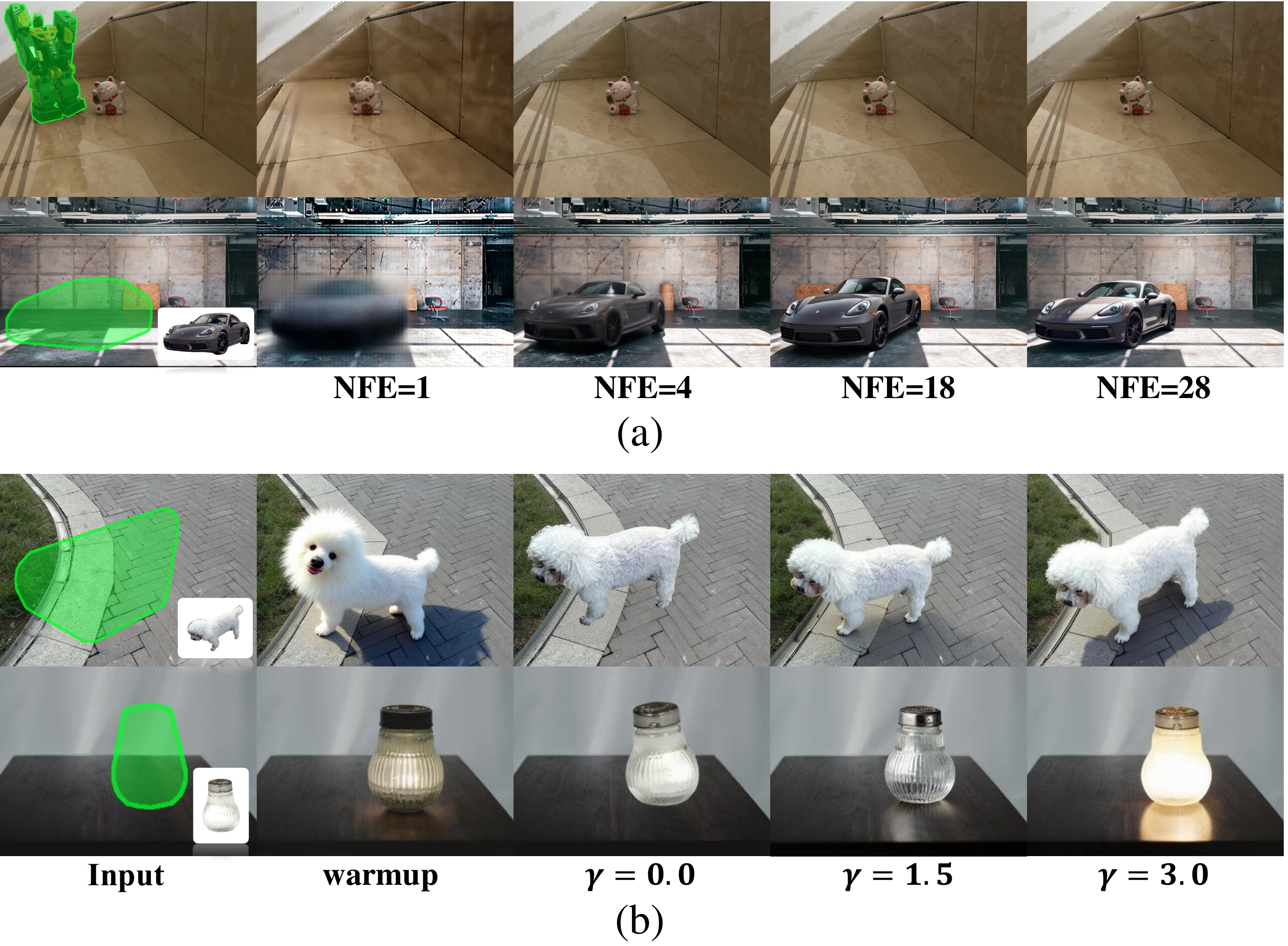}
    \vspace{-7mm}
    \caption{Impact of inference steps and cycle loss weights. \textbf{(a)} Removal (top) and insertion (bottom) results across different neural function evaluations (NFE). \textbf{(b)} Insertion results with varying cycle loss weights $\gamma$, with OmniPaint defaulting to $\gamma=1.5$.}
    \label{fig:ablation_hyperparameters}
    \vspace{-4mm}
\end{figure}

\noindent\textbf{Cycle Loss Weight.}
We analyze the impact of the cycle loss weight $\gamma$ on object insertion by comparing results across different values in Fig.~\ref{fig:ablation_hyperparameters}(b).
Lower $\gamma$ values (e.g., $\gamma=0$) result in weak physical effect synthesis, as the unpaired training data (COCO-Stuff~\cite{COCO-stuff} and HQSeg~\cite{sam_hq}) lack object effects segmentation such as shadows and reflections.
This limits the model's ability to learn effect generation, as insertion training relies on input images that already contain these effects.
Increasing $\gamma$ enhances effect synthesis. At $\gamma=1.5$, OmniPaint achieves the optimal balance, effectively learning from unpaired data while preserving realistic effect synthesis. However, further increasing $\gamma$ to $3.0$ over-relaxes effect generation, leading to unnatural artifacts like exaggerated shadows.

\noindent\textbf{Neural Function Evaluation.}  
We analyze the impact of neural function evaluations (NFE) on object removal and insertion, as illustrated in Fig.~\ref{fig:ablation_hyperparameters}(a). Lower NFE values, such as 1 or 4, lead to noticeable blurring, especially within masked regions. Interestingly, for removal tasks, even NFE=1 effectively eliminates the object and its associated effects. At NFE=18, objects are removed cleanly without residual artifacts, while inserted objects exhibit high fidelity with realistic shading and reflections. Further increasing NFE to 28 yields only marginal gains, indicating diminishing returns. Nonetheless, we set NFE=28 as the default to ensure optimal visual quality.

\section{Conclusion}
We present OmniPaint for object-oriented image editing that reconceptualizes object removal and insertion as interdependent tasks.
By leveraging a pre-trained diffusion prior and a progressive training pipeline comprising initial paired sample optimization and subsequent large-scale unpaired refinement via CycleFlow, OmniPaint achieves precise foreground elimination and seamless object integration while preserving scene geometry and other intrinsic properties. Extensive experiments demonstrate that OmniPaint effectively suppresses object hallucination and mitigates artifacts, with the novel CFD metric providing a robust, reference-free assessment of contextual consistency.

{
    \small
    \bibliographystyle{ieeenat_fullname}
    \bibliography{main}
}

\end{document}

%% file: preamble.tex
\usepackage{times}
\usepackage{epsfig}
\usepackage{graphicx}
\usepackage{amsmath}
\usepackage{amssymb}
\usepackage{algorithm}
\usepackage{algpseudocode}
\usepackage{adjustbox}
\usepackage{multirow,booktabs}
\usepackage{subcaption}
\usepackage{caption}

%% file: sec/fig_teaser.tex
\centering \centering
\includegraphics[width=\textwidth]{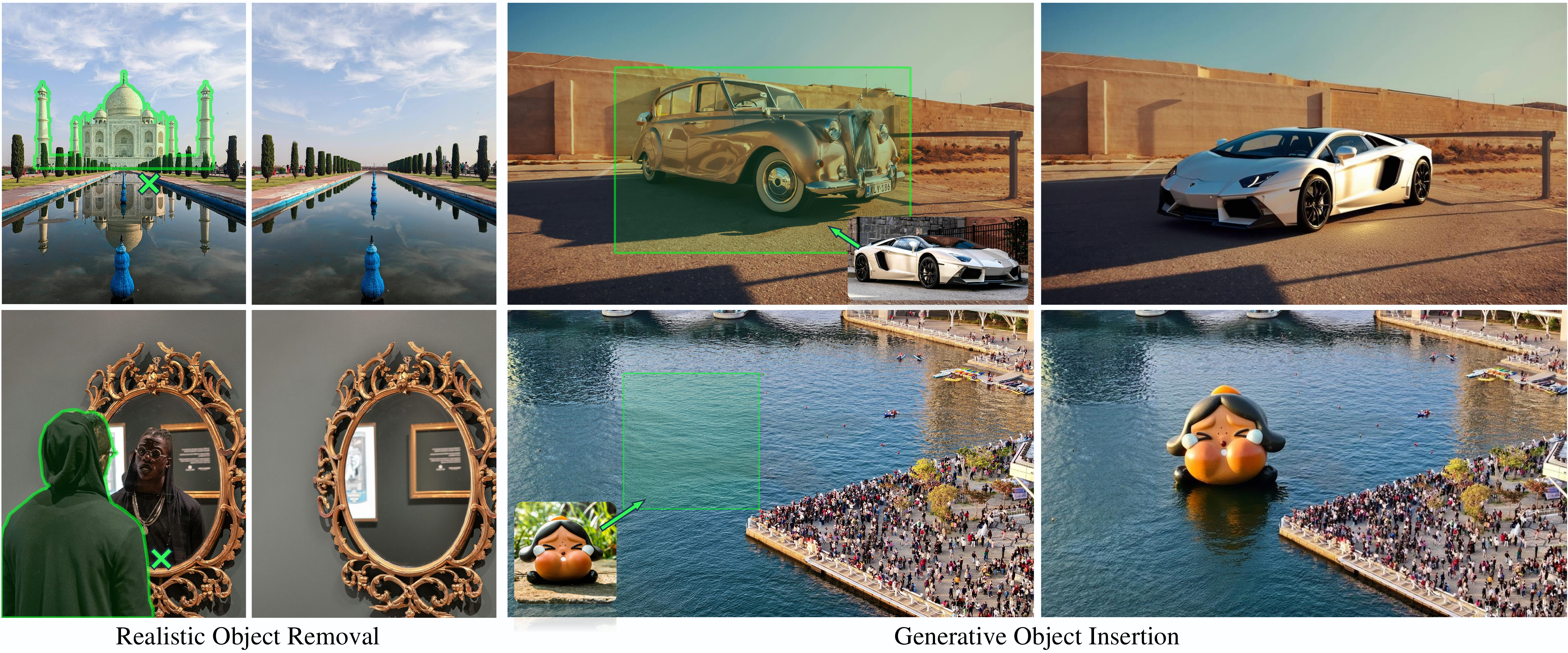}
\captionsetup[figure]{aboveskip=0.05cm}
\captionof{figure}
{
Illustration of OmniPaint for object-oriented editing, including realistic object removal (left) and generative object insertion (right). Masked regions are shown as semi-transparent overlays. In removal cases, the {\color[HTML]{00FF24} $\times$} marks the target object and its physical effects, such as reflections, with the right column showing the results. In insertion cases, the reference object (inset) is placed into the scene, indicated by a green arrow. Note that for model input, masked regions are fully removed rather than semi-transparent.
}
\label{fig:teaser}

%% file: sec/0_abstract.tex
\begin{abstract}
Diffusion-based generative models have revolutionized object-oriented image editing, yet their deployment in realistic object removal and insertion remains hampered by challenges such as the intricate interplay of physical effects and insufficient paired training data. In this work, we introduce OmniPaint, a unified framework that re-conceptualizes object removal and insertion as interdependent processes rather than isolated tasks. Leveraging a pre-trained diffusion prior along with a progressive training pipeline comprising initial paired sample optimization and subsequent large-scale unpaired refinement via CycleFlow, OmniPaint achieves precise foreground elimination and seamless object insertion while faithfully preserving scene geometry and intrinsic properties. Furthermore, our novel CFD metric offers a robust, reference-free evaluation of context consistency and object hallucination, establishing a new benchmark for high-fidelity image editing. Project page: \href{https://yeates.github.io/OmniPaint-Page/}{https://yeates.github.io/OmniPaint-Page/}.
\end{abstract}